\documentclass[conference]{IEEEtran}
\IEEEoverridecommandlockouts
\usepackage{cite}
\usepackage{amsmath,amssymb,amsfonts,amsthm}
\usepackage{algorithmic}
\usepackage{graphicx}
\usepackage{textcomp}
\usepackage{xcolor}
\usepackage{times}
\usepackage{soul}
\usepackage[hidelinks]{hyperref}
\usepackage[utf8]{inputenc}
\usepackage[small]{caption}
\usepackage{graphicx}
\usepackage{booktabs}
\usepackage{subcaption}
\usepackage{color}
\usepackage{hhline}
\theoremstyle{definition}
\newtheorem{definition}{Definition}

\usepackage{bbding}
\usepackage[misc]{ifsym}
\usepackage{url}
\usepackage{bbm}
\usepackage[ruled,linesnumbered]{algorithm2e}
\newcommand\mdoubleplus{\mathbin{+\mkern-10mu+}}

\def\BibTeX{{\rm B\kern-.05em{\sc i\kern-.025em b}\kern-.08em
		T\kern-.1667em\lower.7ex\hbox{E}\kern-.125emX}}
	
\usepackage[absolute,showboxes]{textpos}

\TPMargin{5pt}

\newcommand{\copyrightstatement}{
	\begin{textblock}{0.84}(0.08,0.93)    
		\noindent
		\footnotesize
		\copyright  2019 IEEE. Personal use of this material is permitted. Permission from IEEE must be obtained for all other uses, in any current or future media, including reprinting/republishing this material for advertising or promotional purposes, creating new collective works, for resale or redistribution to servers or lists, or reuse of any copyrighted component of this work in other works.
	\end{textblock}
}

\begin{document}
	\copyrightstatement
	\title{Collaborative Graph Walk for Semi-supervised Multi-Label Node Classification}
	
	\author{\IEEEauthorblockN{Uchenna Akujuobi\IEEEauthorrefmark{1}, Han Yufei\IEEEauthorrefmark{1}\IEEEauthorrefmark{2}, Qiannan Zhang\IEEEauthorrefmark{1}, Xiangliang Zhang\Envelope\IEEEauthorrefmark{1}}
		\IEEEauthorblockA{\IEEEauthorrefmark{1}\textit{King Abdullah University of Science and Technology (KAUST), Saudi Arabia} \\
			\{uchenna.akujuobi, qiannan.zhang, xiangliang.zhang\}@kaust.edu.sa}
		\IEEEauthorblockA{\IEEEauthorrefmark{2}\textit{Symantec, France} \\
			yfhan.hust@gmail.com }}

	\maketitle
	
	\begin{abstract}
		In this work, we study semi-supervised multi-label node classification problem in  attributed graphs. Classic solutions to multi-label node classification follow two steps, first learn node embedding and then build a node classifier on the learned embedding. To improve the discriminating power of the node embedding, we propose a novel collaborative graph walk,   named Multi-Label-Graph-Walk, to finely tune node representations with the available label assignments in   attributed graphs via reinforcement learning. The proposed method formulates the multi-label node classification task as simultaneous graph walks conducted by multiple label-specific agents. Furthermore,  policies of the label-wise graph walks are learned in a cooperative way to capture first the predictive relation between   node labels and    structural attributes of graphs; and second, the correlation among the multiple label-specific classification tasks. A comprehensive experimental study demonstrates that the proposed method can achieve significantly better multi-label classification performance   than the state-of-the-art approaches and conduct more efficient graph exploration.
		
	\end{abstract}
	

	
	
	\section{Introduction}\label{sec:introduction}
	Graph-structured data are frequently witnessed in many real-world applications, such as social graphs and academic graphs. In the graph structure,   nodes  represent   entities (e.g., users in social graphs and papers in citation graphs), whereas   edges linking  two nodes denote the relationship between the entities (e.g., user friendship and paper citation). Usually both   nodes and edges possess their own attributes. For example, a paper node in citation graphs   can present its title/abstract/keywords, while an edge linking two papers has the context of how one paper is cited in the other.   The resultant \emph{attributed  graphs} thus become hosts of rich knowledge about  various real-world physical process and applications. Plenty of research efforts have been devoted on making  comprehensive understanding of attributed graphs \cite{hamilton2017inductive,gcn,yang2016revisiting,lee2018attention}.
	
	In multi-label node classification, the goal is to assign one or more labels to each node. For example,  users of social graphs are   annotated with several tags profiling their preferences in different domains. As well, one paper in  citation networks can be labeled with several  research topics. Multi-label node classification in attributed graphs is a challenging problem on several aspects. First, each node is associated simultaneously with more than one labels, and the  label dependency is difficult to capture in this complex structure.   Multi-label classification models considering the dependency among  labels always perform better than those treating each label independently \cite{wu2014semi,zha2009graph}.  However,  the mixture of node/edge attributes and graph structural information make the label dependency difficult to capture.  
	Second, labels are difficult to obtain. Especially,  the cost of labeling enough nodes   becomes prohibitively expensive as the scale of the networks increases. Moreover, the well known crowd-sourcing issue in multi-label learning makes it even more challenging to provide enough trustable labels to conduct supervised learning for   multi-label classifiers. Therefore, \textbf{semi-supervised multi-label node classification} becomes a  highly demanding, while barely explored technique, where an accurate classifier is trained with a small portion of labeled nodes and plenty of unlabeled nodes. We provide the formal definition of semi-supervised multi-label node classification as follows: 
	\begin{definition}
		\small
		\it
		Semi-supervised Node Classification: Given an attributed graph $G = \{V,E,x_v,x_e\}$, where node set $V$ contains a small subset of labeled nodes $V_l = {<v_i, y_i>},1 \le i \le |V_l|$ and the remaining nodes $ V_u =  V / V_l = {<v_j>}, 1 \le j \le |V_u|$ are unlabeled. $x_v$ and $x_e$ denote the attributes of nodes and edges in the graph $G$, respectively.  The goal is to infer the labels of the unlabeled nodes $V_u$ based on the available while limited node labels, the graph content and structure information. Note that we focus on the multi-label problem, where each   node is associated with multiple labels simultaneously, i.e., $y_i \in 2^{L}$, where  {$L$} is the number of classes.
	\end{definition}
	The most popular solutions to  multi-label node classification follow two steps: 1) learn node embeddings  in an unsupervised way\cite{ grover2016node2vec,perozzi2014deepwalk,tang2015line, nandanwar2016structural,gao2018deep,yang2015network,chen2017hierarchical}; 2) train a multi-label classifier on the learned embeddings. In this process, learning of node embedding is independent of the classification model training. The   embedding is thus not tuned   for the predictive task. Few semi-supervised multi-label classification models are built directly with the original attributes of nodes (not with edges) \cite{wu2014semi, zha2009graph}. 
	However, these models limit the incorporation of pairwise label relevance via   pre-computed metrics. 
	
	In this paper, we propose a \textbf{Multi-Label-Graph-Walk} (MLGW) approach to address semi-supervised multi-label node classification under the framework of reinforcement learning, which aims at integrating information from \textbf{both labeled and unlabeled nodes} into the learning process of node embeddings in attributed graphs. The learned node embeddings are used to conduct both \textbf{transductive} and \textbf{inductive} multi-label node classification. In the proposed MLGW method, we assign an agent for each label and let each label-specific agent walk across the graph and decide which neighboring nodes to explore to win the game (maximizing the classification gain). At each step, the action of each agent moving to the next node is determined by a score network considering the current node attributes, the attributes of the neighboring edges and nodes, and the accumulated history of the previous steps along the path. The resultant walk path can be considered as the agent's recurrent decisions on the informative nodes associated with the target label. 
	
	Our contributions are summarized as follows:
	\begin{itemize}
		\item We cast the problem of semi-supervised multi-label node classification as simultaneous walks of multiple label-specific agents over the attributed graphs. Learning of walk policy for each agent is guided by not only the available but  limited class labels, but also the temporal relations between visited nodes along the walk paths, which encodes node and neighborhood attributes of labeled and unlabeled nodes on the paths.  
		
		\item We propose to formulate the graph walk as Partially-Observed-Markov-Decision-Processes (POMDP) and solve it within the framework of reinforcement learning. The walk policies of each label-specific agent are learned cooperatively to decide what neighborhood information to aggregate and how to plan walk paths across the graph in order to maximize the overall multi-label classification gain. In this sense, the graph walks of different agents are organized as a {\em collaborative exploration} over the multi-label attributed graphs.
		\item Thanks to the learned label-specific policy functions, the proposed model is applicable for both transductive and inductive classification tasks. Extensive experiments on real-world datasets illustrate that our proposed model outperforms the state-of-the-art methods. We also provide case studies to demonstrate the meaningfulness of the guided graph walk.
	\end{itemize}
	
	\section{Previous Work} \label{relatedwork}
	
	\subsection{ Node classification by  embeddings learned via unsupervised methods}
	
	There are unsupervised graph embedding models learning from plain network  \cite{grover2016node2vec,perozzi2014deepwalk,tang2015line}, and attributed network  \cite{gao2018deep,yang2015network,chen2017hierarchical}.
	The  learned embeddings are then given as inputs to multi-class or multi-label classifiers.
	The advantage of the unsupervised node embedding methods is that the learned node embedding is   universally applicable to different tasks (e.g., also for community discovery  and link prediction). The advantage is also the disadvantage, because the node embedding is lack of label information and not particularly tailored for the node classification purpose. 
	
	\subsection{Node classification by  semi-supervised multi-label learning }
	There are semi-supervised multi-class graph embedding models for attributed networks, e.g.,  Planetoid   \cite{yang2016revisiting} GCN \cite{gcn}, DGM \cite{akujuobi2018mining}, GraphSAGE \cite{hamilton2017inductive}, DGCN \cite{zhuang2018dual}, GATs \cite{velickovic2017graph}, LANE \cite{LANE-WSDM17}.
	The node embeddings are learned by incorporating the multi-class label and node content.  These models are designed with the principle that different classes have no overlapping nodes (completely disjoint). Therefore, they are not adoptable to our problem in the multi-label setting.  
	
	To incorporate the label relationship in multi-label problem,   in
	\cite{wu2014semi}, a label-to-label pairwise affinity matrix is used as a regularizer added to the likelihood function of a generative model  estimating the conditional probability of a class observation given an instance. However, enforcing pairwise constraints can be expensive, given a large number of labels. We do not limit our model to   pairwise relationships. In \cite{zha2009graph},   a similar pre-calculated correlation matrix is used in a regularization framework to force the labels learned to be consistent with the observed label correlation. 
	In our proposed model, instead of using a pre-defined label relationship, we let the agents to \textbf{learn} the label relationships during the learning phase.  
	
	\subsection{Reinforcement learning with graph-structured data}
	The use of reinforcement learning on graph-structured data has found applications in different research fields. \cite{xiong2017deeppath} proposed a model to locate multi-hop relationships in knowledge graphs. In the proposed method, an agent locates the informative paths linking entities by incrementally sampling a relation to extend its path. \cite{hoshen2017vain} proposed an attention-based method for learning multi-agent interactions with graph-structured data by applying soft attention to learned agent pair interactions matrix with the motivation of selecting information from relevant agents.  \cite{jiang2018graph} applied multi-head attention to convolutional operations of the agent graph to extract relationship representation between nodes to learn multi-agent cooperation. \cite{lee2017deep} proposed the use of a graph attention walk for multi-class graph classification.
	However, none of these models is designed for multi-label node classification. 
	
	\subsection{Multi-task reinforcement learning}
	Asynchronous reinforcement learning has been introduced to solve the problem of multi-task learning  \cite{METaylor2017AAAI,Tutunov2018NIPS,Calandriello2014NIPS,Mnih2016AsynchronousMF,Espeholt2018,Hessel2019MultitaskDR,Teh2017DistralRM}. Most of the previous works in this area originate from \textit{A3C} \cite{Mnih2016AsynchronousMF}, where an actor is assigned as per task to sample task-specific episodes and a central learner learns a joint policy from all the sampled episodes. 
	The main focus of these methods is to address the bottleneck of large data throughout between local actors and the learner, in order to achieve scalable asynchronous policy exploration \cite{Espeholt2018} and \cite{Hessel2019MultitaskDR}. However, in our study, generating trajectories with positive rewards over the multi-label attributed networks becomes prohibitively expensive, given the increasingly large size of the networks. Recently, a collaborative distributed policy learning method is proposed in \cite{Teh2017DistralRM}, named \textit{Distral}. In this method, each agent learns the task-specific policy locally. Furthermore, a centralized policy is introduced as a global regularization term to guide local policy learning: each local policy is constrained to stay close to the centralized policy. The centralized policy thus improves sampling efficiency of local policy update by capturing and sharing the common knowledge with the local agents. Nevertheless, \textit{Distral} assumes that each agent has the same state and action space. This assumption barely holds in the studied graph walk problem, as historical contexts of the walk and local neighborhoods along the walk path usually vary a lot from one task-specific agent to another. In our work, we borrow the spirit of \textit{Distral} to organize collaborative and efficient policy learning for multi-label node classification.

	\section{Methodology}
	
	\begin{figure*}[h]
		\centering
		\includegraphics[width=0.96\textwidth]{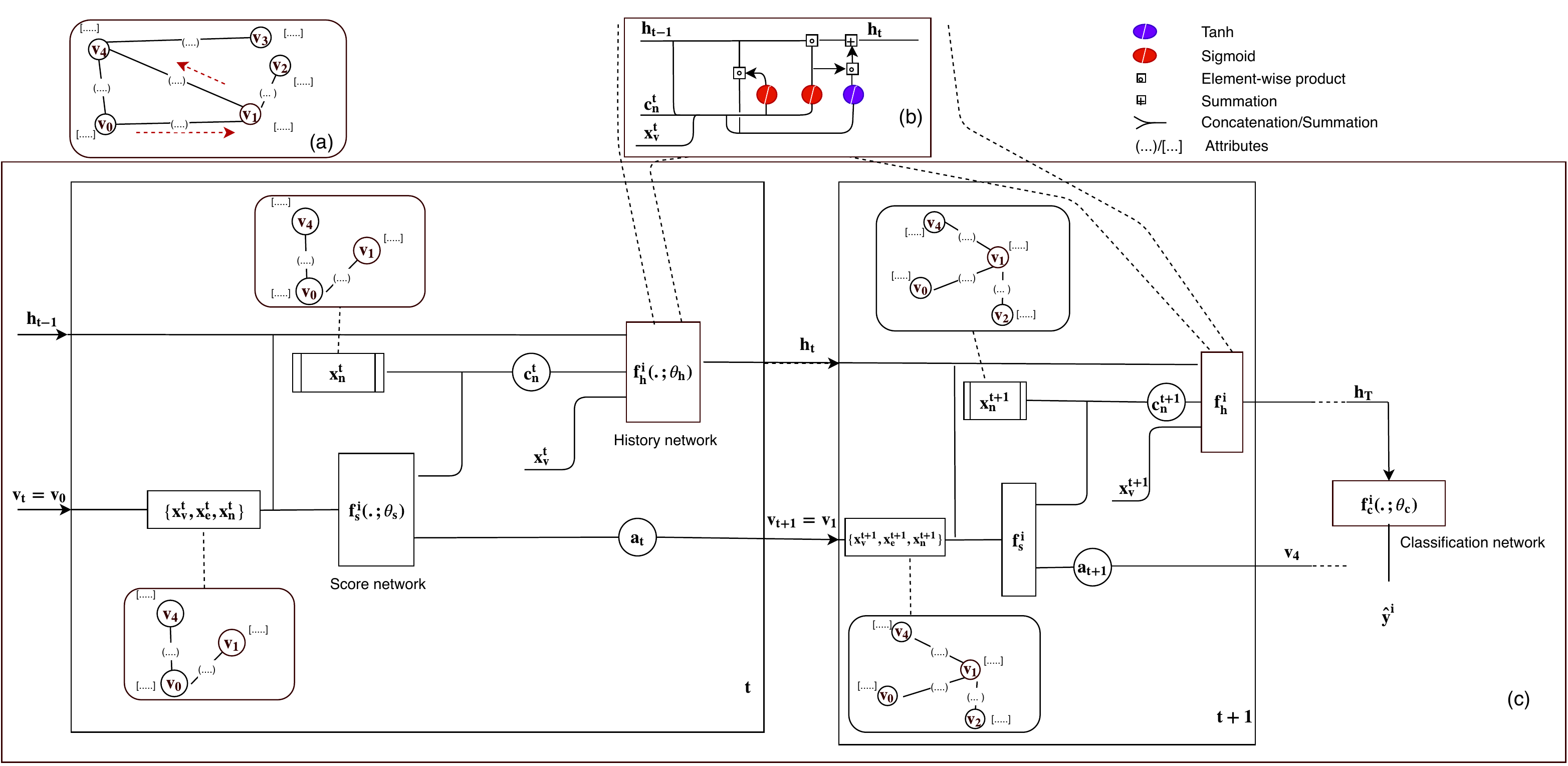}
		\caption{The proposed model. (a) a small attributed graph where the label-specific agent $A_i$ is currently at node $v_0$ and deciding the next visit at time $t+1$, thus $v_t=v_0$. In the left beginning of block (c),  the score network $f^i_s(.;\theta_s)$ takes as input the previous history $h_{t-1}$, the current node attribute $x^t_v$, and the attributes of the immediate node and edge neighbors ${x^t_n, x^t_e}$. The choice $a_t$ of the next node $v_{t+1}$ to visit is sampled from the output of the each score network. Neighboring nodes $x^t_n$ are selected and averaged to form the immediate neighborhood aggregation $c^t_n$. The history network $f^i_h(.;\theta_t)$ takes as input the aggregated neighbor embedding $c^t_n$, the previous history $h_{t-1}$, and the current node embedding $x^t_v$; then outputs the current walk history $h_t$, as shown in (b). At time $t+1$ when the label agent moves to $v_1$,  the same process repeats to move the label agent to $v_4$ at $t+2$, etc. After a number of steps, the final history vector summarizing the information obtained from the graph walk is passed to the classification network $f^i_c(.;\theta_c)$ for classifying the starting node (i.e., deciding if the  node from where $A_i$ started the walk belongs to label $l_i$).}
		\label{fig:full_model}
	\end{figure*}
	
	\subsection{Overview of MLGW}
\begin{figure}[h]
		\centering
		\includegraphics[width=0.56\textwidth]{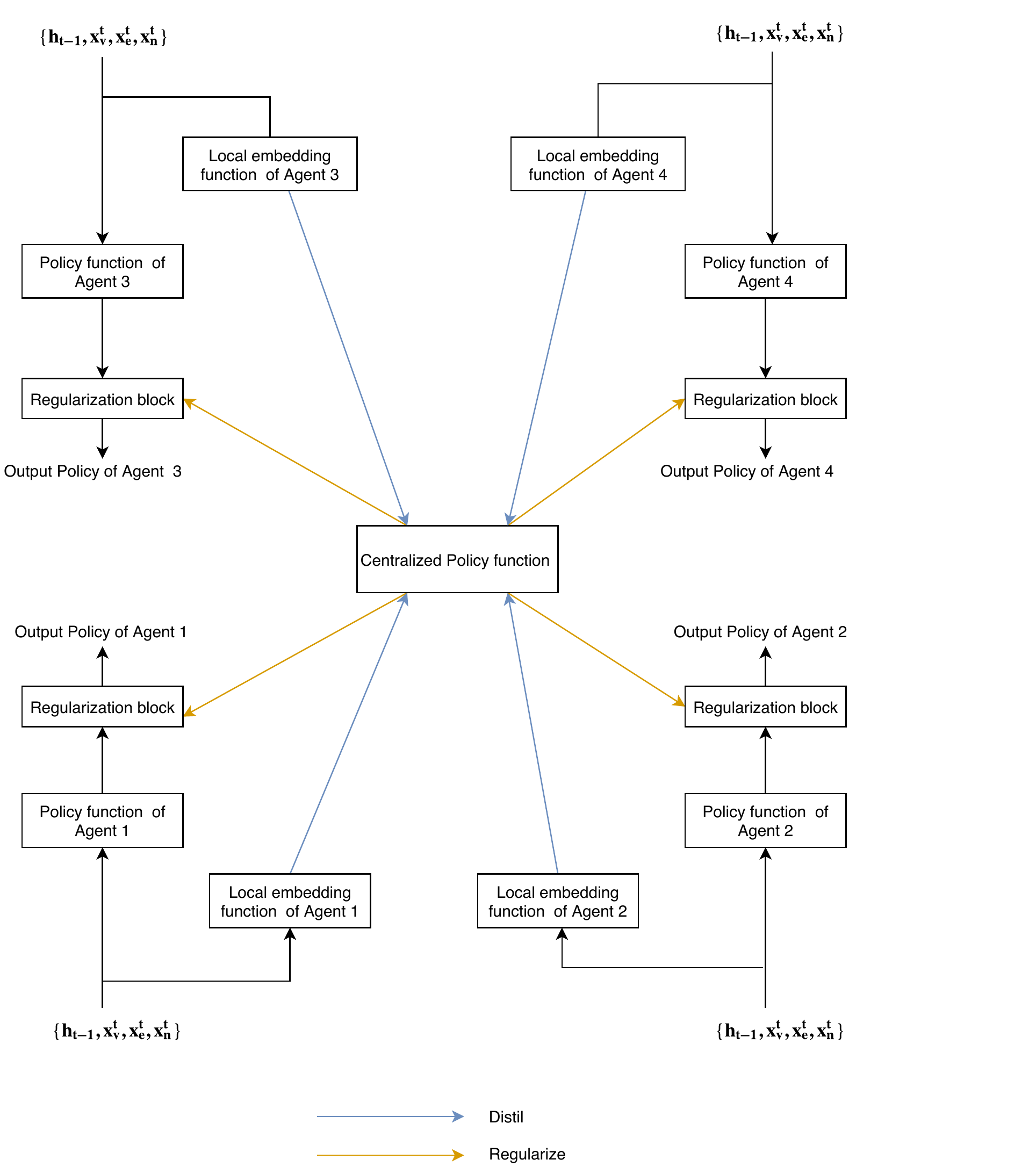}
		\caption{Illustration of the agent communication framework on a network with four possible labels.
			Learning of the centralized policy depends on the historical contexts of the walk path, the embedding of currently visited nodes, the embeddings of all the neighboring nodes and edges from each label-specific agent and the local policy model $p_i$, as shown by Eq.\ref{eq:distilled_pg_global}. 
			The local policy update of each agent takes the regularization   enforced by the centralized policy as defined in Eq.\ref{eq:distilled_pg_local}. 
		}
		\label{fig:agent_com}
	\end{figure}

	As illustrated in Figure \ref{fig:full_model}, in the proposed MLGW model, each label-specific agent explores the attributed networks simultaneously from a starting node. At each step $t$, each agent $A_i$ decides the next node to visit following its stochastic policies. Consequently, the agents develop in parallel multiple paths and update the historical contexts of each walk path recursively. The final historical context information is then used to conduct simultaneous label-specific classification. Such a graph walk process can be formulated as a partially observable Markov decision process (POMDP). To tackle the partial observation at each step, we integrate the walk dependencies recursively on the previous steps and encode the walk sequence by Gated Recurrent Unit (GRU) \cite{cho2014learning}. 
	Therefore, the policy (score) network takes inputs of the current observable environment of the agent and all its past walk path context over the graph. It produces a probabilistic confidence over each node neighbor (potential next nodes to visit). Though the walks are conducted in a label-wise manner, learning of the policies of each agent is organized as a collaborative policy update process, in order to exploit the underlying correlation between label-specific graph walks. We introduce {\em a centralized policy} bridging all the label-specific agents to organize collaborative policy learning of each agent, as shown in Figure \ref{fig:agent_com}. 
	With this design, the policy of each agent is updated based on not only the reward signals received from its walk path but also the simultaneous walk experiences conducted by all the other agents. The resultant policy function, thus by design, encodes the correlation between the label-specific classification tasks. 
	
	\subsection{Components of MLGW }
	Our proposed model consists of three networks for each label specific agent $A_{i}$: the history network $f^i_h(.;\theta_h)$, the score network $f^i_s(.;\theta_s)$, and the classification network $f^i_c(.;\theta_c)$. In Figure \ref{fig:full_model}, our framework snippet for one label-specific agent can be explained as follows: at each step $t$, a label-specific agent $A_i$ is located at node $v_t=v_0$. The history network recursively updates the embedding of past context, where the agent receives the historical context $h_{t-1}$ to define its current status and decide the next move. 
	The decision of where to move at   step $t$ is made upon the policy of $A_i$, which is defined with the score network $f^i_s(.;\theta_s)$, taking as input the 
	history $h_{t-1}$, the current node attribute $x^t_v$, and the attributes of the immediate edge and node neighbors $\{x^t_e,x^t_n\}$.
	This relevance score, ranging from 0 to 1, denotes the relevance of node neighbors to the current node, with 1 signifying a high relevant node neighbor. The choice of the next node to visit is sampled in proportion to their relevance score produced by the score network (see Equation \ref{eq:norm}). Besides, to incorporate the current context,
	an immediate neighborhood information $c^t_n$ is formed by selectively aggregating the embeddings of neighboring nodes $x^t_n$ based on their relevance. 
	The history network $f^i_h(.;\theta_h)$ updates to get history $h_{t}$ as the summary over the past walk path, 
	given as input the previous history $h_{t-1}$, the current node attribute $x^t_v$, and the neighborhood aggregation $c^t_n$. At step $t+1$, the label agent moves to node $v_{t+1}$ and repeats the whole process. After limited walk steps, the final history vector $h_T$, summarizing the walk path of the label agent is passed to the classification network $f^i_c(.;\theta_c)$ to classify the starting node. 
	For a label agent $A_i$, it performs a binary classification  for the label $l_i$ (i.e., belongs to the class or not). 
	The classification process of a given node is described in Algorithm \ref{alg:generl_alg}.
	
	The proposed MLGW model is also applicable in inductive setting,
	where the walk policy is learned based on the nodes available in the graph. Given a new node added to the graph, multiple label agents initiate the walks from the new unlabeled node guided by the learned policies based on their $f^i_s(.;\theta_s)$ and $f^i_h(.;\theta_h)$, and finally use $f^i_c(.;\theta_c)$ for classification. 
	
	\begin{algorithm}[] \small
		\SetAlgoLined
		\SetKwInput{KwIn}{Initialization}
		\KwIn {Graph $G$, start node $v_1$, history vector $h_{0}$ (a vector of zeros)}
		\KwResult{label prediction for node $v_1$}
		
		\For{$t \gets 1 \dotsi T$ }{
			Obtain the     embedding $x^t_v$ of the current node $v_t$; $x^t_e $ for edges connected to $v_t$; and $x^t_n$ for neighboring nodes \;
			
			Assign relevance value to each neighbor node $\varphi^t= f^i_{s}(h_{t-1}, x^t_v, x^t_e, x^t_n; \theta_s)$\;
			
			Sample next node $v_{t+1}$ from the output of the label specific policy $\pi_i = Cat (.|\varphi^t)$ over the neighbors \;
			
			Extract the relevant neighbor information $c^t_n$ \;
			
			update the history vector $h_t = f^i_h(h_{t-1}, x^t_v, c^t_n;\theta_h)$ \;
		}
		
		Obtain the label prediction of the starting node $ y^i_{v_1} = f^i_c(h_T;\theta_c)$

		\caption{Classify node $v_1$ by agent $A_i$ in \textbf{MLGW}}
		\label{alg:generl_alg}
	\end{algorithm}
	
	\subsubsection{\textbf {Information Propagation}}
	Walking over attributed networks is intrinsically a POMDP problem \cite{Shen2018MWalkLT}. Observing only attributes of the neighborhood of a given node is insufficient to differentiate states of a given agent from one to another. It is thus necessary to integrate recurrently the historical information of the graph walk conducted by the agent as an augmented state representation, in order to decide what action to take. The history network computes a history vector $h_t$ over time, which acts as a summary of the walk path. At step $t$, the  network updates the history vector based on the current observation via $h_t = f^i_h(h_{t-1}, x^t_v, c^t_n;\theta_h)$, which has GRU at its core and is formulated as:
	\begin{align}
	z_t = \sigma_g (W^z[x_v^t \mdoubleplus c^t_n] + U^zh_{t-1} + b^z)\nonumber\\
	r_t = \sigma_g (W^r[x_v^t \mdoubleplus c^t_n] + U^rh_{t-1} + b^r)\nonumber\\
	h^\prime_t = \sigma_{h^\prime} (W[x_v^t \mdoubleplus c^t_n]+ r_t\circ Uh_{t-1} + b)\nonumber\\
	h_t = z_t \circ h^\prime_t + (1-z_t) \circ h_{t-1},
	\end{align}
	where $\circ$ and $\mdoubleplus$ denote element-wise multiplication and vector concatenation respectively.  The variable $z_t$ is the update gate which determines the amount of past information to overwrite, $r_t$ is the reset gate which decides the amount of past information to compute a new memory content, $h^\prime_t$ is the current memory content, and $h_t$ is the output vector containing information from the current unit and previous units. The variables $W$ and $U$ are the weights; $x_v^t$ is the node attribute of the current node, $c^t_n$ is the aggregated attribute of the relevant current node neighbors, and $b^z, b^r, b$ are the bias vectors.
	
	At the end of the walk ($t=T$), the history network $f^i_h(.;\theta_h)$ produces $h_T$, the embedding of the full graph walk started from the target node. To classify the target node, $h_T$ is given to the classification network $f^i_c(.;\theta_c)$, modeled as a single-layer neural network, to predict the class label.
	
	\subsubsection{\textbf{Action}}
	\label{action}
	At each step, the label-specific agents make decisions on which node to visit next. This choice is based on the output of the score network $\varphi^t= f^i_{s}(h_{t-1}, x^t_v, x^t_e, x^t_n; \theta_s)$.
	The output $\varphi^t$ of the score network is a measure of relevance of the neighboring nodes to the current node $v_t$. The score network $f^i_s(.;\theta_s)$ is modeled using a single-layer perceptron network with a sigmoid activation function. 
	For better graph exploration, the decision of the next node ($v_{t+1}$) to explore is made via a stochastic policy $\pi_{i}$ by sampling under the categorical distribution $\pi_{i} = Cat(.|\varphi^t)$, after a normalization of the  score values:
	\begin{equation}
	\label{eq:norm}
	\pi_{i} = Cat(.|\varphi^t) = \frac{1}{\sum_{v_k}  \varphi^t_{v_k}} \times \varphi^t_{v_k}; \quad v_k \in N_r(v_t),
	\end{equation}
	where $N_r(v_t)$ is the set of nodes in the one-hop neighborhood of the current node $v_t$,  and $\varphi^t_{v_k} $ is the relevance score of $v_k$.
	
	In our work, we encourage each agent to select the target of the move from the neighbors with the relevance score greater than $0.5$. The aggregation of neighboring information is conducted as: 
	\begin{equation}
	\label{eq:aggregation}
	c_{n}^t = \sum_{v_k} x_{k} \times  \mathbbm{1}(\varphi^t_{v_k} - 0.5), 
	\end{equation}
	where $x_{k}$ is the node attribute of node $v_k$.
	The indicator function $\mathbbm{1}(.)$ outputs 1 when positive and 0 otherwise. 
	
	\subsubsection{\textbf{Reward}}
	We adopt a delayed reward mechanism. Each label-specific agent receives a reward until it reaches to the final classification step at the end of its walk. 
	A label agent receives a reward of $r_t = 1$ at step $T$  for a correct final node classification and $r_t = -1$ otherwise. 
	
	\subsection{Collaborative Policy Learning with Centralized Policy based Regularization}\label{subsec:train}
	As unveiled in \cite{Yu:2014}, the key to successful multi-label classification is to capture the underlying correlation between simultaneous label-specific classification tasks. Following this spirit, we impose a centralized policy as a global constraint in the policy learning process of each agent. This design is inspired by \cite{Teh2017DistralRM}, which is designated to capture correlation among task-specific knowledge. As we will discuss later in this section, the centralized policy modifies the policy update steps, so that policy learning of each label-specific agent is conducted by considering policies of all the other agents. We denote the label-specific policy as $\pi_{i}$ and the distilled policy as $\pi_{d}$. The joint policy learning objective gives as in Eq. (\ref{eq:distilled_policy}): 
	\begin{equation}\label{eq:distilled_policy}
	\begin{split} \small
	&\mathcal{J}(\{\theta_{i=1}^{n}\},\theta_{d}) = \sum_{i}E_{\pi_{i}}[\sum_{t\geq0}(\gamma^{T-t}r^{i} \\
	&-\alpha \gamma^{T-t}\log\frac{\pi_{i}(a^{i}_{t}|s^{i}_{1:t},\theta_{i})}{\pi_{d}(a^{i}_{1:t}|s^{i}_{1:t},\theta_{i})}-\beta\gamma^{T-t}\log\pi_{i}(a^{i}_t|s^{i}_{1:t},\theta_{i}))],\\
	\end{split}
	\end{equation}
	where $s^{i}_{1:t}$ is the historical summary of the walk path conducted by the agent of the label $l_i$ following its policy $\pi_{i}$. $a^{i}_t$ denotes the walk decision (e.g., move to the next neighboring nodes) taken by the agent of the label $l_i$ at a given step $t$ in the interaction sequence. $r^i$ is the delayed reward that the agent $i$ receive at the end of each walk path. $\gamma \in (0, 1]$ is a discount factor to emphasize the decisions made near the classification. $T$ is the length of the walk path. 
	$\alpha$ and $\beta$ are weight parameters determining the strengths of \textbf{KL-divergence} and \textbf{entropy-based} regularization terms. $\{\theta_{i=1,2,...,L}\}$ and $ \theta_{d}$ denotes the parameters of the label-specific policy and the global distilled policy function. On one hand, the KL-divergence term regularizes the output of each $\pi_{i}$ towards the distilled policy $\pi_{d}$. On the other hand, the entropy regularization is employed to encourage policy exploration and avoid to be trapped in local optimum of policy exploration. 
	
	It is, however, non-trivial to optimize Eq. (\ref{eq:distilled_policy}) directly with respect to the joint distribution of all possible interaction experiences. We thus adopt REINFORCE \cite{williams1992simple}, a policy gradient method, to address the optimization problem in Eq. (\ref{eq:distilled_policy}). Proposed initially to solve MDPs, policy gradient has also been widely known as an attractive and scalable approach for controlling POMDP \cite{williams1992simple}. We discretize the learning objective with sampled interaction sequences and derive the gradient with respect to the parameters $\theta_{i}$ and $\theta_{d}$ in each label-specific policy function, given as follows: 
	
	\begin{equation}\label{eq:distilled_pg_local}
	\small
	\begin{split}
	&\nabla_{\theta_{i}} \mathcal{J} = \\
	&\frac{1}{N}\sum_{m=1}^{N}\sum^{T}_{t=1} \nabla_{\theta_{i}} \log \pi_{i} (a^{i}_{m,t} |\theta_{i},s^{i}_{m,1:T}) \left(\sum_{u\geq{t}}{\gamma^{T-u}}\hat{R}_{i}(a^{i}_{m,u},s^{i}_{m,1:u})\right)\\
	&\hat{R}_{i}(a^{i}_{m,u},s^{i}_{m,1:u}) = r_{i} + \frac{\tilde{\alpha}}{\tilde{\beta}}\log\pi_{d}(a^{i}_{m,u}|\theta_{d},s^{i}_{m,1:u}) \\
	&- \frac{1}{\tilde{\beta}}\log \pi_{i}(a^{i}_{m,u}|\theta_{i},s^{i}_{m,1:u})\\
	\end{split}
	\end{equation}
	\begin{equation}\label{eq:distilled_pg_global}
	\small
	\begin{split}
	&\nabla_{\theta_{d}} \mathcal{J} = \frac{1}{N}\sum_{i=1}^{L}\sum_{m=1}^{N}\{\\
	&\sum^{T}_{t=1} \nabla_{\theta_{d}} \log \pi_{i} (a^{i}_{m,t} |\theta_{i},s^{i}_{m,1:T}) \left(\sum_{u\geq{t}}{\gamma^{T-u}}\hat{R}_{i}(a^{i}_{m,u},s^{i}_{m,1:u})\right)\\
	+&\frac{\tilde{\alpha}}{\tilde{\beta}}\sum_{t=1}^{T}\gamma^{T-t}(\pi_{i}(a^{i}_{m,t})-\pi_{d}(a^i_{m,t}))\nabla_{\theta_{d}} \log \pi_{d} (a^{i}_{m,t} |\theta_{d},s^{i}_{m,1:t})\},\\
	\end{split}
	\end{equation}
	where $\tilde{\alpha} = \frac{\alpha}{\alpha+\beta}$ and $\tilde{\beta}=\frac{1}{\alpha+\beta}$.  $\hat{R}_{i}(a^{i}_{m,u},s^{i}_{m,1:u})$ is the regularizer on the reward received by the label-specific agent. $N$ is the number of the sampled sequences by each agent. As we can find, the update of the global distilled policy matches the probabilities under the task policy $\pi_{i}$ and under the distilled policy $\pi_{d}$.\textcolor{blue}{} The KL-divergence based regularizor forces the global policy to be the centroid of all label-specific policies, which helps transfer knowledge about policies of graph walk across different label-specific agents.
	It is worth noting that the centralized policy $\pi_{d}$ only serves as a global regularization in the policy gradient. 
	Nevertheless, it does not prevent from incorporating the output from both label-specific policies and the centralized policy together to decide the walk path. Though there is no theoretical preference over either way of using $\pi_{d}$ organizing graph walk, we will evaluate and compare empirically the performances of both strategies in Section \ref{subsec:node_classification}.
	
	For the classification network, we use a Hybrid supervised loss \cite{mnih2014recurrent}: we train the classification network $f^i_{c}(.;\theta_c)$ for each label. We use the cross-entropy based loss function to maximizes the conditional likelihood of the true label $\log \pi_{i}(l_i | s_{1: T}; \theta_c )$, given the observations from the trajectory of graph walk $ s_{1: T}$, where $l_i$ is the given true label. 
	
	\section{Experiments}
	\subsection{Datasets}
	\label{dataset_des}
	
	The datasets used in our experiments are multi-labeled datasets extracted from the DBLP and Delve databases. The DBLP (four area) dataset \cite{ji2010graph} is a multi-label citation dataset. We construct a co-authorship graph where each node represents an author, and the edge signifies a co-authorship. 
	The label set is the research areas: database (DB - {\it ID = 0}),  data mining (DM - {\it ID = 1}), artificial intelligence (AI - {\it ID = 2}) and information retrieval (IR - {\it ID = 3}). The node attributes are the concatenated titles of the papers published by the author. The DBLP dataset has no edge attribute. We describe how we deal with missing edge attributes in section \ref{comparison_and_setup}.
	
	The Delve dataset\footnote{Extracted from http://delve.kaust.edu.sa} is a multi-label citation dataset where each node represents a paper, and the edges show the citation relationship between papers. The Delve dataset label set consists of 20 predefined topics in machine learning and data-mining (see Table \ref{delve_label}). The node attributes are the title and abstract (when available) of the papers, and the edge attributes are the citation contexts, i.e., the sentences encompassing the citations. We evaluate on two versions of the dataset: Delve-M \cite{akujuobi2018mining}, and Delve-R (extended to include more papers). We show the dataset statistics in Table \ref{dataset_info}.
	
	\begin{table}[ht!]
		\caption{Statistics of datasets used in  evaluations,  showing the number of nodes $|V|$, number of edges $|E|$, number of labels $|L|$, number of labeled nodes $|V^L|$, and averge label cardinality $\hat{C}$. }
		\centering
		\footnotesize 
		\begin{tabular}{|l|rrrrr|}
			\hline
			& $\mid V\mid$ & $\mid E\mid$   & $\mid L\mid$   & $\mid V^L\mid$ & $\hat{C}$    \\ \hline
			DBLP  & 28,702     & 68,335   & 4         & 28,702 &  1.18 \\ \hline
			Delve-M     & 1,229,280 & 4,322,275 & 20       & 3,686  & 1.25\\ \hline
			Delve-R    & 1,229,280 & 4,322,275 & 20       & 131,991  & 1.2\\ \hline
		\end{tabular}
		\label{dataset_info}
	\end{table}
	
	\begin{table*}[]
		\centering
		\caption{Evaluation results on the DBLP dataset.}
		\label{dblp}
		\begin{tabular}{lrrrrrrrrrrrr}
			\hline
			\multicolumn{13}{c}{DBLP}                                                                                                                                                                                                                                                                                                                                                                             \\ \hline
			\multicolumn{1}{l|}{}                    & \multicolumn{6}{c|}{Tr-1}                                                                                                                                                        & \multicolumn{6}{c}{Tr-4}                                                                                                                                                \\ \hline
			\multicolumn{1}{l|}{}                    & \multicolumn{2}{c}{precision}                          & \multicolumn{2}{c|}{Recall}                            & \multicolumn{2}{c|}{F1}                                        & \multicolumn{2}{c|}{precision}                         & \multicolumn{2}{c|}{Recall}                            & \multicolumn{2}{c}{F1}                                \\ \hline
			\multicolumn{1}{l|}{}                    & \multicolumn{1}{l}{macro} & \multicolumn{1}{l|}{micro} & \multicolumn{1}{l}{macro} & \multicolumn{1}{l|}{micro} & \multicolumn{1}{l}{macro} & \multicolumn{1}{l|}{micro}         & \multicolumn{1}{l}{macro} & \multicolumn{1}{l|}{micro} & \multicolumn{1}{l}{macro} & \multicolumn{1}{l|}{micro} & \multicolumn{1}{l}{macro} & \multicolumn{1}{l}{micro} \\ \hline
			\multicolumn{1}{l|}{BR}                  & 79.9                      & \multicolumn{1}{r|}{81.1}  & 70.0                      & \multicolumn{1}{r|}{72.4}  & 74.5                      & \multicolumn{1}{r|}{76.5}          & 82.7                      & \multicolumn{1}{r|}{83.7}  & 70.7                      & \multicolumn{1}{r|}{73.2}  & 76.0                      & 78.1                      \\
			\multicolumn{1}{l|}{LP}                  & 77.7                      & \multicolumn{1}{r|}{79.1}  & 70.5                      & \multicolumn{1}{r|}{73.4}  & 73.7                      & \multicolumn{1}{r|}{76.1}          & 81.7                      & \multicolumn{1}{r|}{82.6}  & 71.9                      & \multicolumn{1}{r|}{74.9}  & 76.2                      & 78.5                      \\
			\multicolumn{1}{l|}{CC}                  & 76.8                      & \multicolumn{1}{r|}{77.8}  & 7.02                      & \multicolumn{1}{r|}{74.5}  & 74.2                      & \multicolumn{1}{r|}{76.1}          & 79.4                      & \multicolumn{1}{r|}{80.1}  & 73.6                      & \multicolumn{1}{r|}{76.3}  & 76.2                      & 78.2                      \\
			\multicolumn{1}{l|}{Rk}                  & 77.7                      & \multicolumn{1}{r|}{79.1}  & 70.5                      & \multicolumn{1}{r|}{73.4}  & 73.7                      & \multicolumn{1}{r|}{76.1}          & 81.7                      & \multicolumn{1}{r|}{82.6}  & 71.9                      & \multicolumn{1}{r|}{74.9}  & 76.2                      & 78.5                      \\
			\multicolumn{1}{l|}{MLKNN}               & 68.3                      & \multicolumn{1}{r|}{70.7}  & 60.7                      & \multicolumn{1}{r|}{63.5}  & 64.2                      & \multicolumn{1}{r|}{66.9}          & 80.7                      & \multicolumn{1}{r|}{82.0}  & 75.8                      & \multicolumn{1}{r|}{77.3}  & 78.1                      & 79.6                      \\
			\multicolumn{1}{l|}{MARM}                & 55.6                      & \multicolumn{1}{r|}{52.1}  & 55.8                      & \multicolumn{1}{r|}{62.8}  & 48.2                      & \multicolumn{1}{r|}{56.9}          & 62.7                      & \multicolumn{1}{r|}{58.2}  & 65.6                      & \multicolumn{1}{r|}{69.7}  & 58.1                      & 63.4                      \\
			\multicolumn{1}{l|}{GF}                  & 70.7                      & \multicolumn{1}{r|}{73.0}  & 62.4                      & \multicolumn{1}{r|}{65.5}  & 66.1                      & \multicolumn{1}{r|}{69.0}          & 82.7                      & \multicolumn{1}{r|}{83.8}  & 76.7                      & \multicolumn{1}{r|}{78.4}  & 79.6                      & 81.0                      \\
			\multicolumn{1}{l|}{GraphSAGE\_mean}     & 73.7                      & \multicolumn{1}{r|}{75.9}  & 73.5                      & \multicolumn{1}{r|}{75.1}  & 73.5                      & \multicolumn{1}{r|}{75.5}          & 83.9                      & \multicolumn{1}{r|}{85.1}  & 84.5                      & \multicolumn{1}{r|}{85.6}  & 84.1                      & 85.3                      \\
			\multicolumn{1}{l|}{GraphSAGE\_GCN}      & 78.3                      & \multicolumn{1}{r|}{79.8}  & 65.6                      & \multicolumn{1}{r|}{67.5}  & 71.2                      & \multicolumn{1}{r|}{73.1}          & 82.1                      & \multicolumn{1}{r|}{83.7}  & 80.2                      & \multicolumn{1}{r|}{81.7}  & 81.1                      & 82.7                      \\
			\multicolumn{1}{l|}{GraphSAGE\_maxpool}  & 76.8                      & \multicolumn{1}{r|}{78.6}  & 71.6                      & \multicolumn{1}{r|}{73.3}  & 74.1                      & \multicolumn{1}{r|}{75.8}          & 84.4                      & \multicolumn{1}{r|}{85.8}  & 85.6                      & \multicolumn{1}{r|}{86.6}  & 85.0                      & 86.2                      \\
			\multicolumn{1}{l|}{GraphSAGE\_meanpool} & 72.2                      & \multicolumn{1}{r|}{74.4}  & 73.4                      & \multicolumn{1}{r|}{74.4}  & 72.4                      & \multicolumn{1}{r|}{74.4}          & 83.6                      & \multicolumn{1}{r|}{85.0}  & 85.3                      & \multicolumn{1}{r|}{86.3}  & 84.4                      & 85.6                      \\
			\multicolumn{1}{l|}{GraphSAGE\_LSTM}     & 70.5                      & \multicolumn{1}{r|}{73.2}  & 73.8                      & \multicolumn{1}{r|}{75.4}  & 71.9                      & \multicolumn{1}{r|}{74.3}          & 84.0                      & \multicolumn{1}{r|}{85.3}  & 84.5                      & \multicolumn{1}{r|}{85.6}  & 84.2                      & 85.4                      \\
			\multicolumn{1}{l|}{MLGW-I\_TRANS}       & 80.4                      & \multicolumn{1}{r|}{81.7}  & 74.0                      & \multicolumn{1}{r|}{76.2}  & 76.8                      & \multicolumn{1}{r|}{78.8}          & 87.1                      & \multicolumn{1}{r|}{87.9}  & 83.8                      & \multicolumn{1}{r|}{85.3}  & 85.4                      & 86.6                      \\
			\multicolumn{1}{l|}{MLGW-I\_IND}         & 80.4                      & \multicolumn{1}{r|}{81.5}  & 73.9                      & \multicolumn{1}{r|}{76.3}  & 76.7                      & \multicolumn{1}{r|}{78.8}          & 87.0                      & \multicolumn{1}{r|}{88.0}  & 83.6                      & \multicolumn{1}{r|}{85.0}  & 85.2                      & 86.5                      \\
			\multicolumn{1}{l|}{MLGW-REG\_TRANS}     & 81.0                      & \multicolumn{1}{r|}{82.5}  & 74.4                      & \multicolumn{1}{r|}{76.7}  & \textbf{77.4}             & \multicolumn{1}{r|}{\textbf{79.5}} & 88.4                      & \multicolumn{1}{r|}{89.2}  & 86.4                      & \multicolumn{1}{r|}{87.6}  & 87.3                      & 88.4                      \\
			\multicolumn{1}{l|}{MLGW-REG\_IND}       & 81.0                      & \multicolumn{1}{r|}{82.4}  & 74.5                      & \multicolumn{1}{r|}{76.9}  & \textbf{77.4}             & \multicolumn{1}{r|}{\textbf{79.6}} & 88.2                      & \multicolumn{1}{r|}{89.1}  & 86.3                      & \multicolumn{1}{r|}{87.5}  & 87.2                      & 88.3                      \\
			\multicolumn{1}{l|}{MLGW-REG+\_TRANS}    & 78.6                      & \multicolumn{1}{r|}{80.0}  & 73.3                      & \multicolumn{1}{r|}{74.9}  & 75.8                      & \multicolumn{1}{r|}{77.4}          & 89.5                      & \multicolumn{1}{r|}{90.3}  & 87.5                      & \multicolumn{1}{r|}{88.5}  & \textbf{88.5}             & \textbf{89.4}             \\
			\multicolumn{1}{l|}{MLGW-REG+\_IND}      & 79.0                      & \multicolumn{1}{r|}{80.3}  & 72.5                      & \multicolumn{1}{r|}{74.3}  & 75.4                      & \multicolumn{1}{r|}{77.2}          & 89.9                      & \multicolumn{1}{r|}{90.7}  & 87.2                      & \multicolumn{1}{r|}{88.2}  & \textbf{88.5}             & \textbf{89.4}             \\ \hline
		\end{tabular}
	\end{table*}

	\begin{table*}[]
		\centering
		\caption{Evaluation results on the Delve-M database}
		\label{delve_m}
		\begin{tabular}{lrrrrrrrrrrrr}
			\hline
			\multicolumn{13}{c}{Delve-M}                                                                                                                                                                                                                                                                                                                                                                          \\ \hline
			\multicolumn{1}{c|}{}                    & \multicolumn{6}{c|}{Tr-1}                                                                                                                                                        & \multicolumn{6}{c}{Tr-4}                                                                                                                                                \\ \hline
			\multicolumn{1}{c|}{}                    & \multicolumn{2}{c|}{precision}                         & \multicolumn{2}{c|}{Recall}                            & \multicolumn{2}{c|}{F1}                                        & \multicolumn{2}{c|}{precision}                         & \multicolumn{2}{c|}{Recall}                            & \multicolumn{2}{c}{F1}                                \\ \hline
			\multicolumn{1}{l|}{}                    & \multicolumn{1}{c}{macro} & \multicolumn{1}{c|}{micro} & \multicolumn{1}{c}{macro} & \multicolumn{1}{c|}{micro} & \multicolumn{1}{c}{macro} & \multicolumn{1}{c|}{micro}         & \multicolumn{1}{c}{macro} & \multicolumn{1}{c|}{micro} & \multicolumn{1}{c}{macro} & \multicolumn{1}{c|}{micro} & \multicolumn{1}{c}{macro} & \multicolumn{1}{c}{micro} \\ \hline
			\multicolumn{1}{l|}{BR}                  & 57.9                      & \multicolumn{1}{r|}{64.0}  & 36.2                      & \multicolumn{1}{r|}{48.7}  & 42.5                      & \multicolumn{1}{r|}{55.3}          & 63.8                      & \multicolumn{1}{r|}{68.2}  & 46.9                      & \multicolumn{1}{r|}{56.2}  & 53.0                      & 61.6                      \\
			\multicolumn{1}{l|}{LP}                  & 47.0                      & \multicolumn{1}{r|}{55.6}  & 36.5                      & \multicolumn{1}{r|}{48.8}  & 40.1                      & \multicolumn{1}{r|}{52.0}          & 53.0                      & \multicolumn{1}{r|}{60.9}  & 42.0                      & \multicolumn{1}{r|}{53.8}  & 45.7                      & 57.1                      \\
			\multicolumn{1}{l|}{CC}                  & 57.9                      & \multicolumn{1}{r|}{62}    & 37.3                      & \multicolumn{1}{r|}{49.7}  & 42.9                      & \multicolumn{1}{r|}{55.2}          & 60.8                      & \multicolumn{1}{r|}{66.3}  & 48.7                      & \multicolumn{1}{r|}{58.2}  & 53.3                      & 62.0                      \\
			\multicolumn{1}{l|}{Rk}                  & 47.0                      & \multicolumn{1}{r|}{55.6}  & 36.5                      & \multicolumn{1}{r|}{48.8}  & 40.1                      & \multicolumn{1}{r|}{52.0}          & 53.0                      & \multicolumn{1}{r|}{60.9}  & 42.0                      & \multicolumn{1}{r|}{53.8}  & 45.7                      & 57.1                      \\
			\multicolumn{1}{l|}{MLKNN}               & 50.4                      & \multicolumn{1}{r|}{59.9}  & 28.1                      & \multicolumn{1}{r|}{39.3}  & 33.4                      & \multicolumn{1}{r|}{47.4}          & 58.3                      & \multicolumn{1}{r|}{62.3}  & 34.1                      & \multicolumn{1}{r|}{42.9}  & 40.9                      & 50.8                      \\
			\multicolumn{1}{l|}{MARM}                & 7.2                       & \multicolumn{1}{r|}{17.5}  & 17.3                      & \multicolumn{1}{r|}{38.1}  & 7.2                       & \multicolumn{1}{r|}{23.3}          & 9.1                       & \multicolumn{1}{r|}{22.6}  & 13.9                      & \multicolumn{1}{r|}{32.7}  & 8.3                       & 26.0                      \\
			\multicolumn{1}{l|}{GF}                  & 49.4                      & \multicolumn{1}{r|}{60.0}  & 28.0                      & \multicolumn{1}{r|}{39.3}  & 33.3                      & \multicolumn{1}{r|}{47.4}          & 58.4                      & \multicolumn{1}{r|}{62.4}  & 34.2                      & \multicolumn{1}{r|}{43.0}  & 41.0                      & 50.9                      \\
			\multicolumn{1}{l|}{GraphSAGE\_mean}     & 5.4                       & \multicolumn{1}{r|}{14.2}  & 7.5                       & \multicolumn{1}{r|}{26.9}  & 4.1                       & \multicolumn{1}{r|}{18.3}          & 14.1                      & \multicolumn{1}{r|}{26.4}  & 29.0                      & \multicolumn{1}{r|}{55.9}  & 16.2                      & 35.6                      \\
			\multicolumn{1}{l|}{GraphSAGE\_GCN}      & 5.6                       & \multicolumn{1}{r|}{14.1}  & 7.7                       & \multicolumn{1}{r|}{27.9}  & 4.6                       & \multicolumn{1}{r|}{18.4}          & 9.3                       & \multicolumn{1}{r|}{20.3}  & 28.3                      & \multicolumn{1}{r|}{64.8}  & 12.1                      & 30.9                      \\
			\multicolumn{1}{l|}{GraphSAGE\_maxpool}  & 2.6                       & \multicolumn{1}{r|}{7.9}   & 6.4                       & \multicolumn{1}{r|}{36.3}  & 2.1                       & \multicolumn{1}{r|}{11.1}          & 1.5                       & \multicolumn{1}{r|}{3.9}   & 6.3                       & \multicolumn{1}{r|}{54.3}  & 1.7                       & 6.7                       \\
			\multicolumn{1}{l|}{GraphSAGE\_meanpool} & 3.9                       & \multicolumn{1}{r|}{12.7}  & 6.1                       & \multicolumn{1}{r|}{27.1}  & 2.0                       & \multicolumn{1}{r|}{15.5}          & 0.8                       & \multicolumn{1}{r|}{1.6}   & 6.6                       & \multicolumn{1}{r|}{53.8}  & 1.3                       & 3.2                       \\
			\multicolumn{1}{l|}{GraphSAGE\_LSTM}     & 1.5                       & \multicolumn{1}{r|}{2.5}   & 5.1                       & \multicolumn{1}{r|}{25}    & 1.2                       & \multicolumn{1}{r|}{3.9}           & 2.9                       & \multicolumn{1}{r|}{8.3}   & 9.9                       & \multicolumn{1}{r|}{65.5}  & 3.8                       & 14.5                      \\
			\multicolumn{1}{l|}{MLGW-I\_TRANS}       & 63.8                      & \multicolumn{1}{r|}{71.1}  & 42.5                      & \multicolumn{1}{r|}{56.1}  & 49.1                      & \multicolumn{1}{r|}{62.6}          & 63.7                      & \multicolumn{1}{r|}{69.0}  & 55.4                      & \multicolumn{1}{r|}{64.7}  & 58.1                      & 66.8                      \\
			\multicolumn{1}{l|}{MLGW-I\_IND}         & 63.9                      & \multicolumn{1}{r|}{71.2}  & 42.4                      & \multicolumn{1}{r|}{55.9}  & 49.0                      & \multicolumn{1}{r|}{62.6}          & 63.3                      & \multicolumn{1}{r|}{69.2}  & 55.3                      & \multicolumn{1}{r|}{64.9}  & 58.0                      & 67.0                      \\
			\multicolumn{1}{l|}{MLGW-REG\_TRANS}     & 62.6                      & \multicolumn{1}{r|}{70.6}  & 43.5                      & \multicolumn{1}{r|}{56.2}  & 49.7                      & \multicolumn{1}{r|}{62.5}          & 63.5                      & \multicolumn{1}{r|}{68.2}  & 55.7                      & \multicolumn{1}{r|}{65.3}  & 58.4                      & 66.7                      \\
			\multicolumn{1}{l|}{MLGW-REG\_IND}       & 62.3                      & \multicolumn{1}{r|}{70.4}  & 43.6                      & \multicolumn{1}{r|}{56.3}  & 49.7                      & \multicolumn{1}{r|}{62.5}          & 63.8                      & \multicolumn{1}{r|}{68.3}  & 55.9                      & \multicolumn{1}{r|}{65.4}  & 58.6                      & 66.8                      \\
			\multicolumn{1}{l|}{MLGW-REG+\_TRANS}    & 66.2                      & \multicolumn{1}{r|}{71.0}  & 43.2                      & \multicolumn{1}{r|}{56.3}  & \textbf{50.1}             & \multicolumn{1}{r|}{\textbf{62.7}} & 64.5                      & \multicolumn{1}{r|}{68.9}  & 55.6                      & \multicolumn{1}{r|}{65.2}  & 58.6                      & 67.0                      \\
			\multicolumn{1}{l|}{MLGW-REG+\_IND}      & 66.3                      & \multicolumn{1}{r|}{71.3}  & 43.2                      & \multicolumn{1}{r|}{56.0}  & \textbf{50.1}             & \multicolumn{1}{r|}{\textbf{62.7}} & 64.7                      & \multicolumn{1}{r|}{69.0}  & 56.1                      & \multicolumn{1}{r|}{65.7}  & \textbf{59.0}             & \textbf{67.3}             \\ \hline
		\end{tabular}
	\end{table*}

	\begin{table*}[]
		\centering
		\caption{Evaluation results on the Delve-R dataset}
		\label{delve_r}
		\begin{tabular}{lrrrrrrrrrrrr}
			\hline
			\multicolumn{13}{c}{Delve-R}                                                                                                                                                                                                                                                                                                                                                                          \\ \hline
			\multicolumn{1}{c|}{}                    & \multicolumn{6}{c|}{Tr-1}                                                                                                                                                        & \multicolumn{6}{c}{Tr-4}                                                                                                                                               \\ \hline
			\multicolumn{1}{c|}{}                    & \multicolumn{2}{c|}{precision}                         & \multicolumn{2}{c|}{Recall}                            & \multicolumn{2}{c|}{F1}                                        & \multicolumn{2}{c|}{precision}                         & \multicolumn{2}{c|}{Recall}                            & \multicolumn{2}{c}{F1}                               \\ \hline
			\multicolumn{1}{l|}{}                    & \multicolumn{1}{l}{macro} & \multicolumn{1}{l|}{micro} & \multicolumn{1}{l}{macro} & \multicolumn{1}{l|}{micro} & \multicolumn{1}{l}{macro} & \multicolumn{1}{l|}{micro}         & \multicolumn{1}{l}{macro} & \multicolumn{1}{l|}{micro} & \multicolumn{1}{l}{macro} & \multicolumn{1}{l|}{micro} & \multicolumn{1}{l}{macro} & \multicolumn{1}{l}{micro} \\ \hline
			\multicolumn{1}{l|}{BR}                  & 80.9                      & \multicolumn{1}{r|}{87.8}  & 66.5                      & \multicolumn{1}{r|}{76.6}  & 72.5                      & \multicolumn{1}{r|}{81.8}          & 83.9                      & \multicolumn{1}{r|}{89.5}  & 67.8                      & \multicolumn{1}{r|}{77.4}  & 74.1                      & 83.0                      \\
			\multicolumn{1}{l|}{LP}                  & 75.1                      & \multicolumn{1}{r|}{83.3}  & 64.2                      & \multicolumn{1}{r|}{75.0}  & 69.0                      & \multicolumn{1}{r|}{78.9}          & 80.4                      & \multicolumn{1}{r|}{87.0}  & 67.5                      & \multicolumn{1}{r|}{77.4}  & 73.0                      & 81.9                      \\
			\multicolumn{1}{l|}{CC}                  & 78.8                      & \multicolumn{1}{r|}{86.3}  & 68.5                      & \multicolumn{1}{r|}{78.0}  & 72.9                      & \multicolumn{1}{r|}{81.9}          & 81.6                      & \multicolumn{1}{r|}{88.2}  & 70.5                      & \multicolumn{1}{r|}{79.1}  & 74.9                      & 83.4                      \\
			\multicolumn{1}{l|}{Rk}                  & 75.1                      & \multicolumn{1}{r|}{83.3}  & 64.2                      & \multicolumn{1}{r|}{75.0}  & 69.0                      & \multicolumn{1}{r|}{78.9}          & 80.4                      & \multicolumn{1}{r|}{87.0}  & 67.5                      & \multicolumn{1}{r|}{77.4}  & 73.0                      & 81.9                      \\
			\multicolumn{1}{l|}{MLKNN}               & 65.6                      & \multicolumn{1}{r|}{74.0}  & 46.2                      & \multicolumn{1}{r|}{58.1}  & 53.2                      & \multicolumn{1}{r|}{65.1}          & -                         & \multicolumn{1}{r|}{-}     & -                         & \multicolumn{1}{r|}{-}     & -                         & -                         \\
			\multicolumn{1}{l|}{MARM}                & 9.4                       & \multicolumn{1}{r|}{22.7}  & 5.3                       & \multicolumn{1}{r|}{19}    & 2.4                       & \multicolumn{1}{r|}{20.7}          & 13.4                      & \multicolumn{1}{r|}{23.4}  & 5.7                       & \multicolumn{1}{r|}{19.9}  & 3.0                       & 21.5                      \\
			\multicolumn{1}{l|}{GF}                  & \multicolumn{1}{r}{65.7}  & \multicolumn{1}{r|}{74.1}  & \multicolumn{1}{r}{46.3}  & \multicolumn{1}{r|}{58.2}  & \multicolumn{1}{r}{53.3}  & \multicolumn{1}{r|}{65.2}          & \multicolumn{1}{r}{67.5}  & \multicolumn{1}{r|}{75.4}  & \multicolumn{1}{r}{50.9}  & \multicolumn{1}{r|}{62.1}  & \multicolumn{1}{r}{57.4}  & \multicolumn{1}{r}{68.1}  \\
			\multicolumn{1}{l|}{GraphSAGE\_mean}     & 59.6                      & \multicolumn{1}{r|}{72.0}  & 74.6                      & \multicolumn{1}{r|}{81.3}  & 65.2                      & \multicolumn{1}{r|}{76.4}          & 68.6                      & \multicolumn{1}{r|}{78.3}  & 74.7                      & \multicolumn{1}{r|}{81.9}  & 71.3                      & 80.1                      \\
			\multicolumn{1}{l|}{GraphSAGE\_GCN}      & 51.3                      & \multicolumn{1}{r|}{62.6}  & 66.5                      & \multicolumn{1}{r|}{75.4}  & 57.3                      & \multicolumn{1}{r|}{68.4}          & 56.5                      & \multicolumn{1}{r|}{66.8}  & 69.8                      & \multicolumn{1}{r|}{78.2}  & 62.0                      & 72.1                      \\
			\multicolumn{1}{l|}{GraphSAGE\_maxpool}  & 60.0                      & \multicolumn{1}{r|}{72.0}  & 77.0                      & \multicolumn{1}{r|}{83.4}  & 66.4                      & \multicolumn{1}{r|}{77.3}          & 70.2                      & \multicolumn{1}{r|}{79.3}  & 74.9                      & \multicolumn{1}{r|}{82.4}  & 72.3                      & 80.8                      \\
			\multicolumn{1}{l|}{GraphSAGE\_meanpool} & 59.4                      & \multicolumn{1}{r|}{71.2}  & 76.4                      & \multicolumn{1}{r|}{82.9}  & 65.9                      & \multicolumn{1}{r|}{76.6}          & 69.2                      & \multicolumn{1}{r|}{78.6}  & 75.1                      & \multicolumn{1}{r|}{82.6}  & 71.9                      & 80.5                      \\
			\multicolumn{1}{l|}{GraphSAGE\_LSTM}     & 58.1                      & \multicolumn{1}{r|}{69.9}  & 75.4                      & \multicolumn{1}{r|}{82.4}  & 64.9                      & \multicolumn{1}{r|}{75.6}          & 67.1                      & \multicolumn{1}{r|}{77.1}  & 74.2                      & \multicolumn{1}{r|}{81.3}  & 70.1                      & 79.1                      \\
			\multicolumn{1}{l|}{MLGW-I\_TRANS}       & 78.0                      & \multicolumn{1}{r|}{85.5}  & 75.0                      & \multicolumn{1}{r|}{84.3}  & 76.0                      & \multicolumn{1}{r|}{84.9}          & 80.8                      & \multicolumn{1}{r|}{87.0}  & 78.5                      & \multicolumn{1}{r|}{86.6}  & 79.4                      & 86.8                      \\
			\multicolumn{1}{l|}{MLGW-I\_IND}         & 78.0                      & \multicolumn{1}{r|}{85.5}  & 74.9                      & \multicolumn{1}{r|}{84.3}  & 76.0                      & \multicolumn{1}{r|}{84.9}          & 80.7                      & \multicolumn{1}{r|}{86.9}  & 78.9                      & \multicolumn{1}{r|}{86.7}  & \textbf{79.6}             & 86.8                      \\
			\multicolumn{1}{l|}{MLGW-REG\_TRANS}     & 78.5                      & \multicolumn{1}{r|}{85.9}  & 74.9                      & \multicolumn{1}{r|}{84.1}  & \textbf{76.2}             & \multicolumn{1}{r|}{\textbf{85.0}} & 80.0                      & \multicolumn{1}{r|}{85.9}  & 74.6                      & \multicolumn{1}{r|}{83.0}  & 76.9                      & 84.4                      \\
			\multicolumn{1}{l|}{MLGW-REG\_IND}       & 78.3                      & \multicolumn{1}{r|}{85.6}  & 74.9                      & \multicolumn{1}{r|}{84.2}  & \textbf{76.2}             & \multicolumn{1}{r|}{84.9}          & 80.0                      & \multicolumn{1}{r|}{85.9}  & 74.4                      & \multicolumn{1}{r|}{82.9}  & 76.8                      & 84.3                      \\
			\multicolumn{1}{l|}{MLGW-REG+\_TRANS}    & 77.9                      & \multicolumn{1}{r|}{85.6}  & 74.8                      & \multicolumn{1}{r|}{84.2}  & 75.9                      & \multicolumn{1}{r|}{84.9}          & 81.0                      & \multicolumn{1}{r|}{87.1}  & 78.4                      & \multicolumn{1}{r|}{86.6}  & 79.4                      & \textbf{86.9}             \\
			\multicolumn{1}{l|}{MLGW-REG+\_IND}      & 78.2                      & \multicolumn{1}{r|}{85.7}  & 74.4                      & \multicolumn{1}{r|}{84.1}  & 75.8                      & \multicolumn{1}{r|}{84.9}          & 81.0                      & \multicolumn{1}{r|}{87.1}  & 78.4                      & \multicolumn{1}{r|}{86.6}  & 79.4                      & \textbf{86.9}             \\ \hline
		\end{tabular}
	\end{table*}

	\subsection{Comparison methods and experimental setup}
	\label{comparison_and_setup}
	To evaluate the performance of the proposed model, we compare it against several state-of-the-art multi-label classification methods. To avoid unfair comparison due to  faulty implementation,  we evaluate against methods available using the scikit-multilearn library or from author provided codes. The baseline  methods can be split into two groups:

	\paragraph{Unsupervised embedding + multi-label classifier}
	
	We generate embedding using several unsupervised methods for attributed and unattributed graphs (i.e., Deepwalk, Doc2vec, TADW). However, due to space constraints, we report results on using Node2vec and LSI to capture the graph structure and text information as it achieved the best results. The concatenation of resultant embedding vectors 
	are then used as input for several multi-learning classification methods:
	\begin{itemize}
		\item SVM\footnote{SVM is selected because it outperforms logistic regression on all our evaluation cases.} extended by strategies  including binary relevance ({\bf BR}), label powerset ({\bf LP}), classifier chain ({\bf CC}), and RakelD ({\bf Rk}) \cite{zhang2014review,5567103};
		\item MLkNN,   a k-nn adaptation for multi-label tasks \cite{zhang2007ml}; 
		\item Graph factorization   ({\bf GF}) \cite{szymanski2018lnemlc};
		\item  ART classifier ({\bf MARM})\cite{7395756}.
	\end{itemize}

	\paragraph{Graph-based Multi-label method}
	
	We compare against several variants of the supervised GraphSAGE method \cite{hamilton2017inductive}, an inductive method for multi-label node classification. Note that our proposed method works in both transductive and inductive settings. Semi-supervised    models such as GCN, Planetoid, and GAT  are not considered as baselines because  they are designed for multi-class, not   for multi-label problems.  
	
	When the abstract is available, a paper (node) attribute is given as a concatenation of both the title and abstract else only the title is used. Each citation relationship (edge) attribute is given as the concatenation of all its citation contexts (i.e., sentences where the reference is mentioned in the citing paper). The paper and citation attributes are then converted to a vector by applying the latent semantic analysis (LSI) method on the document-term matrix features, resulting in 300-dimension features vectors. We complete the missing citation attributes with zero vectors and assume no missing paper attribute. In all the experiments, the attribute vector is normalized to unit norm.  
	
	For our proposed model, we performed a grid search over the walk length $T=\{5,10,20,40\}$ and the number of walks per node $M=\{1,3,5\}$ (since we train the whole model in parallel end-to-end, we are limited in the 
	number of walks per node due to GPU memory constraints). For each neural network based model, we performed a grid search over the learning rate $lr=\{1e^{-2},5e^{-2},1e^{-3},5e^{-4},1e^{-4}\}$ and hidden layer dimension $d=\{32,64,128\}$. We performed the parameter grid search by training on the DBLP dataset using 20\% labeled samples. The best parameters per model from the grid search are then used in all experiments. 
	We used the default parameters for the SVM strategies, Graph factorization, and ART classifier. For the MLkNN method, we set $k=3$ from a grid search on $k = \{1, 3,5,7\}$.
	The GraphSAGE models are trained for 40 epochs with a parameter set of ($d=128,lr=1e^{-2}$).
	Our proposed method is trained for 20 epochs with a parameter set of  ($d=128,T=10,lr=1e^{-2},M=3, \gamma=0.9, \alpha=1, \beta=0.1$). All reported results are obtained from 5-fold cross-validation using a stratified iterative splitting algorithm \cite{sechidis2011stratification} to ensure the availability of all labels in each fold. We report the results obtained by training on 1-fold and test on 4 folds ({\bf Tr-1}), as well as training 4 folds and testing on 1-fold ({\bf Tr-4}).
	The experiments reported in this paper are performed on a Linux system with an NVIDIA GTX1080Ti GPU. Our model implementation\footnote{https://github.com/Uchman21/MLGW} was conducted by Python using the Tensorflow library.
	
	
	\begin{figure}[t!]
		\scriptsize
		\centering
		\includegraphics[width=0.5\textwidth]{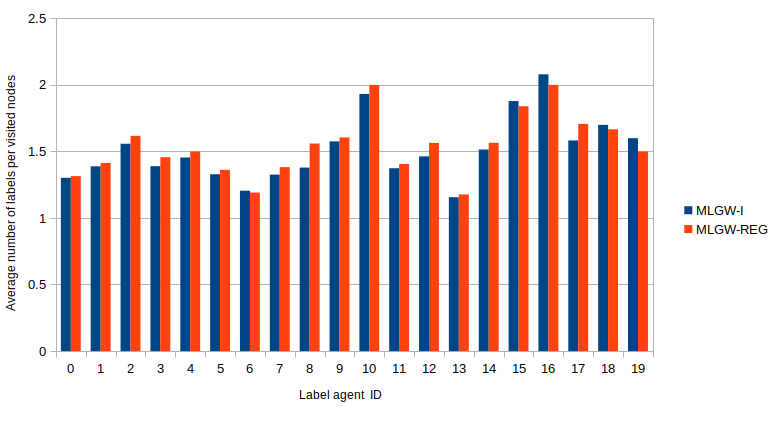}
		\caption{The average number of labels per visited node by each label agent during the graph walk}
		\label{lpvn}
	\end{figure}
	
	\begin{figure*}
		\centering
		\begin{subfigure}[b]{0.49\textwidth}
			\includegraphics[width=0.7\textwidth]{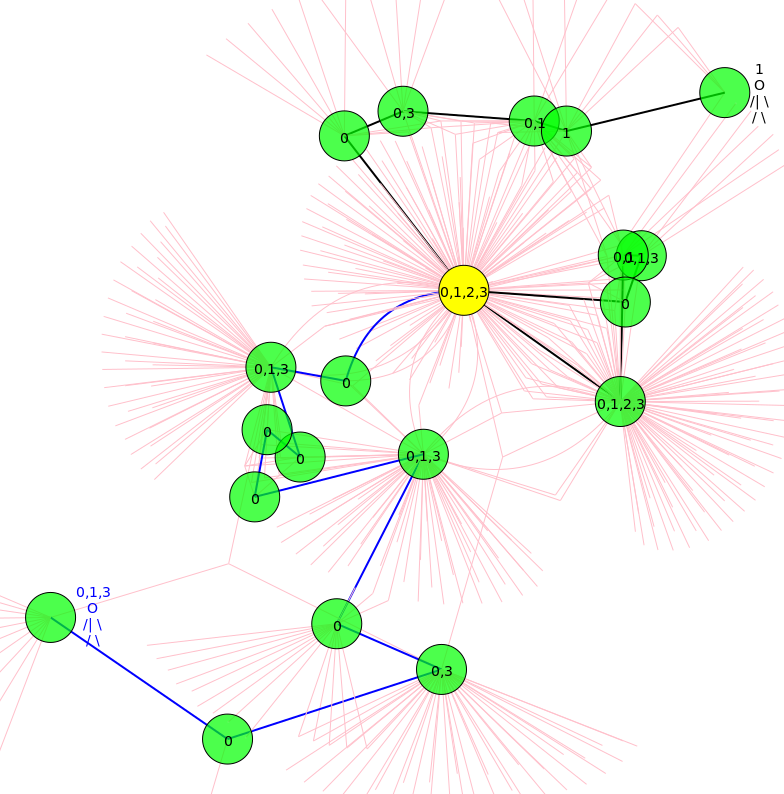}
			\caption{Independent walks}
			\label{fig:ind_walk}
		\end{subfigure}
		~ 
		\begin{subfigure}[b]{0.49\textwidth}
			\includegraphics[width=0.6\textwidth]{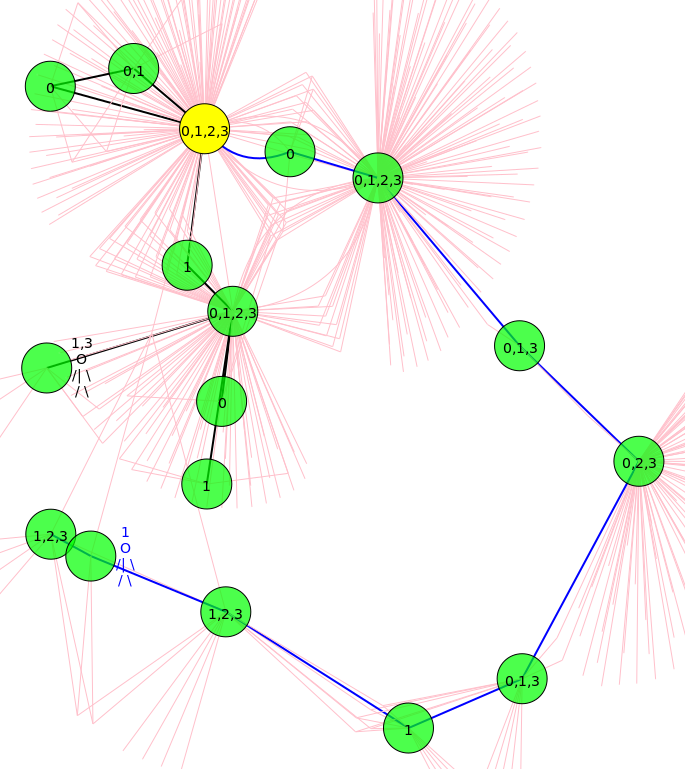}
			\caption{Walk with global policy}
			\label{fig:kl_walk}
		\end{subfigure}
		\caption{Subgraphs showing trajectories of two label agents (for label 0 and 1) using the same settings starting from the yellow node, with labels $\{0,1,2,3\}$ and terminating at the nodes with the stick figures. The black trajectory is of the label agent 1, and the blue trajectory is of label agent 0. Both explore nodes with labels that belong to the starting node, indicated in green color. Label  IDs are shown in the DBLP dataset description (see section \ref{dataset_des}).}\label{fig:graphwalk}
	\end{figure*}

	\subsection{Node Classification Results and Analysis}\label{subsec:node_classification}
	
	We show the performance of our proposed method against several baseline methods in Tables \ref{dblp}, \ref{delve_m}, and \ref{delve_r}. The empty entries in Table \ref{delve_r} denote unavailable results due to scalability issues. Our model outperforms the baseline methods on all datasets. We also evaluate several variants of our model in both {\it inductive (IND)} and {\it transductive (TRANS)} setting to evaluate its advantage on different aspects:
	\begin{itemize}
		\item MLGW independent walks ({\bf MLGW-I}): In this variant, we let the agents make independent walks on the graph without a global policy regularization. Thus, no information sharing among the agents.
		\item MLGW for regularization ({\bf MLGW-REG}): This variant uses the global policy for regularization in the cost function, but makes decisions on which node to move using the local policy output. 
		\item MLGW for regularization and decision ({\bf MLGW-REG+}): This variant uses the global policy for regularization in the cost function and also for the decisions in the graph walk. That is, the joint policy used in deciding the next node to visit for agent $A_i$ is $\pi^{joint}_i = \pi_i \pi_d$.
	\end{itemize}

	We made several interesting observations from the result of the experiments. First, the introduction of a global policy improved the predictive performance in our datasets. However, the improvement is only marginal compared to our proposed independent walk variant. We analyze the results obtained and attribute the marginal improvement to the nature of our graph datasets. In our datasets, a node can have more than one label. Hence, during the graph walk, the label agents can capture the label dependencies even in the independent graph walk setting due to the exploration of nodes with multiple labels. In Figure \ref{fig:graphwalk}, we observe that in walking with or without a global policy, the agent can discover and explore similar neighboring nodes as it is trained to explore relevant nodes to improve the classification task. However, in a graph-structured data with multi-labeled nodes, there could be several paths with relevant nodes. To better capture the label inter-dependencies, walks with global policy tend to favor the exploration of nodes with multiple labels.
	
	This phenomenon can be observed in Figure \ref{fig:graphwalk} as well as in Figure \ref{lpvn}. In Figure \ref{lpvn}, we show the average number of labels per node visited during the graph walk of the label agents trained on the Delve-M dataset with and without the global policy. It can be observed that overall, the label agents trained with the inclusion of the global policy tend to explore nodes with a higher number of labels. This observation shows that the global policy encourages the exploration of nodes that will further capture the inter-dependencies between the labels. For instance, observing a node with more than one label gives more information about the label relationships than observing a node with a single label. The label agents for {\it ``Feature selection \& extraction'', ``Unsupervised learning'', and ``Dimensionality reduction''} explored nodes with higher number of labels. This observation is logical as these three topics are often studied with several other topics, and thus, the nodes explored tend to have more than one label.
	
	\begin{figure}[t]
		\scriptsize
		\centering
		\includegraphics[width=0.45\textwidth]{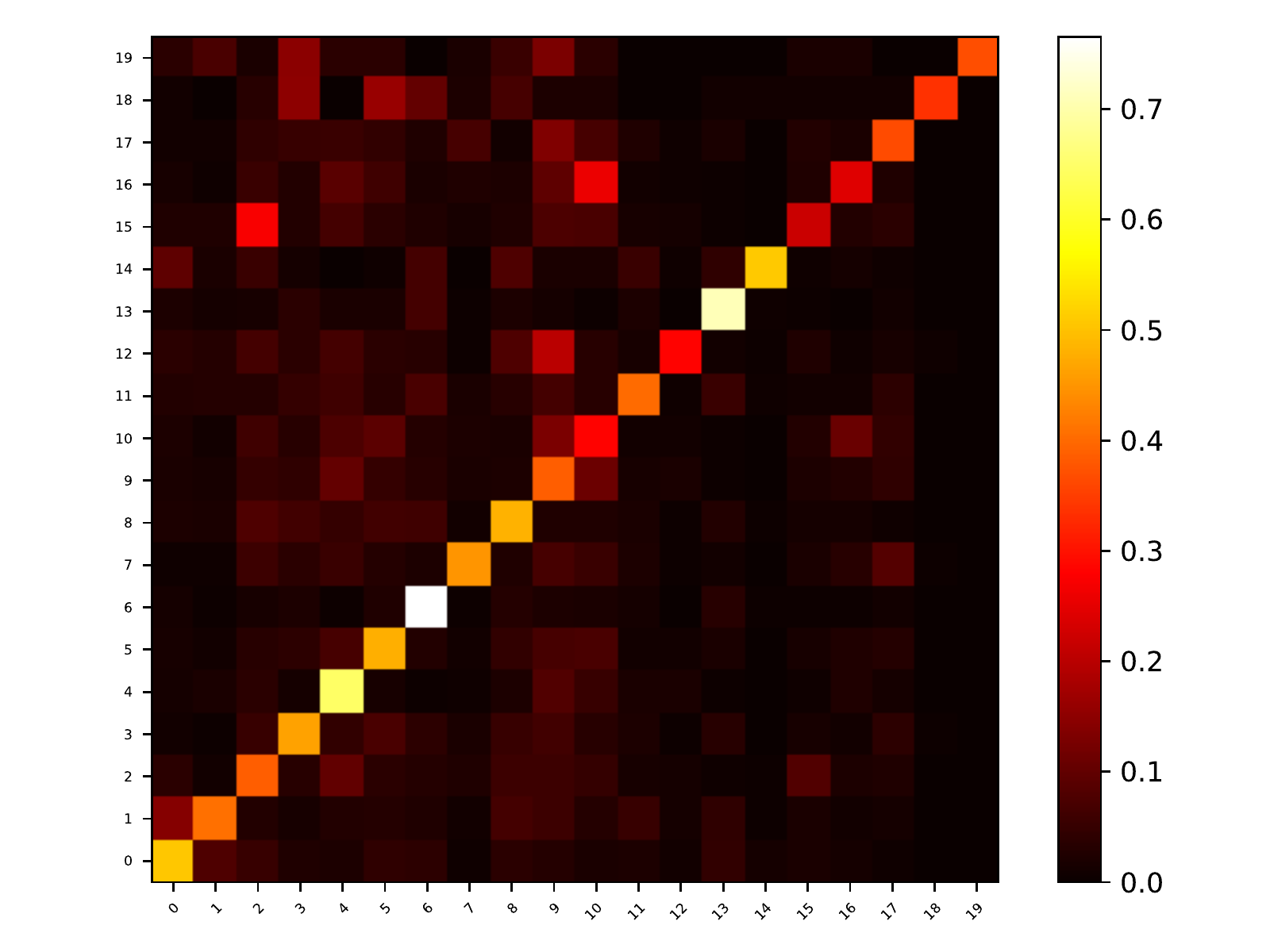}
		\caption{A heatmap whose $d$-th column demonstrates the label distribution of nodes visited by the MLGW-REG+ label agent $d$ starting from nodes with label $d$. The visiting frequency rate is shown in color. A brighter color indicates more frequent visits. It is worth mentioning that agents have no information about any label when walking, neither the label of starting node nor the label on neighboring nodes}
		\label{case_study}
	\end{figure}

	\begin{table}[t]
		\caption{Delve Label-ID Mapping }
		\label{delve_label}
		\centering
		\begin{tabular}{|l|l|l|l|} 
			\hline
			\textbf{Label} & \textbf{ID} & \textbf{Label} & \textbf{ID}\\
			\hline
			Information retrieval & 0 &NLP& 1  \\
			\hline
			Clustering & 2 & Optimization methods & 3\\
			\hline
			Bioinformatics & 4 & Computer vision & 5\\
			\hline
			Security and privacy & 6 & Time series & 7\\
			\hline
			Graph mining \& social network & 8 & Supervised learning & 9\\
			\hline
			Feature selection \& extraction & 10 & Rule learning & 11\\
			\hline
			Semisupervised \& active learning & 12 & Agent systems (AI) & 13\\
			\hline
			Recommendation & 14 & Unsupervised learning & 15\\
			\hline
			Dimensionality reduction & 16 & Neural networks & 17\\
			\hline
			Online learning & 18 & Multi-label classification & 19\\
			\hline
			
		\end{tabular}
	\end{table}
	
	Tables \ref{dblp} \ref{delve_m}, and \ref{delve_r} also show the comparison of inductive MLGW. In the inductive setting, the testing nodes are removed from the training graph and thus are not seen during training. The  optimal policy learned by agents for the graph walk during training   will be generalized to unseen nodes. During testing, the previously unseen nodes are then added to the graph. The agents (guided by the learned policy), start walks from the added new nodes to learn their embedding and predict their labels. We compare against the supervised variants of GraphSAGE that do not use the unlabeled and test nodes during training. The superior performance of MLGW shows that walks starting from the new nodes guided by the learned policy aggregated the most useful information for the classification purpose.

	\begin{figure}[t]
		\scriptsize
		\centering
		\includegraphics[width=0.46\textwidth]{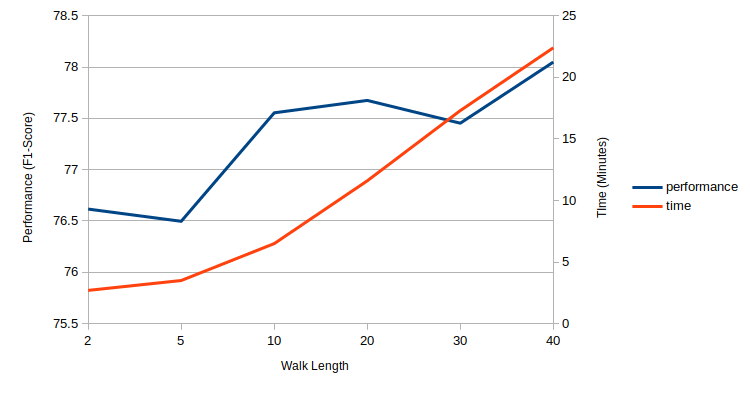}
		\caption{Impact of the walk length on the predictive performance and   time cost}
		\label{parameter}
	\end{figure}

	\subsection{Trajectory Analysis}
	To understand the decision process, 
	we analyze the trajectories learned by the agents assigned to each label. To this end, we trained our model using the full DBLP dataset. For each label (agent), we extract all trajectories starting at papers belonging to the given label. We then analyze the labels of the nodes traversed in each trajectory set. 
	The motivation behind this evaluation is to understand how the label agents determine what nodes to visit for accumulating information for predicting the label of the starting node.
	Figure \ref{fig:graphwalk} shows the trajectories of two label agents with the same settings. 
	We see that the agents can make a guided walk visiting mostly nodes with the same label as the starting node (shown in green).
	Figure \ref{case_study} shows a heatmap. The $d$-th column of the heatmap demonstrates, what types of nodes the agent visited after starting from nodes with $d$-th label. We report the result observed on our trained model using the full Delve-M dataset. 
	With the listed topic IDs in Table \ref{delve_label}, we can see in column with ID=2 (\emph{clustering}), the brightest unit corresponds to ID=15 (\emph{unsupervised learning}).
	In column with ID=10 (\emph{feature selection}), the brightest unit corresponds to ID=16 (\emph{dimensionality reduction}). 
	Interesting observations from this figure verify that agents move to nodes that can help in predicting the labels of the starting nodes. 
	By moving on the attributed network, agents finally learned the intrinsic label relationship among labels.

	\subsection{Parameter study}
	We study the impact of the walk length $T=\{2,5,10,20,30,40\}$ on the performance of the model. We train the model on the DBLP dataset with 20\% training samples and 80\% testing samples. In Figure \ref{parameter}, it can be observed that the model already performs well after ten steps.   However, walking with more steps   results in higher time cost.
	
	\section{Conclusion}
	In this paper, we introduce MLGW, a reinforcement learning based semi-supervised node classification method, for multi-labeled graphs with node and edge attributes. We pose the classification task in partially observed Markov decision processes of multiple label-specific agents. The classification procedure consists of simultaneous recurrent walks conducted by the label-specific agents across the attributed graph. The learning procedure of the policies of each agent is conducted via on-policy reinforcement learning, in order to estimate the parameters of the policy functions maximizing the expected overall classification performances. Furthermore, we introduce a centralized policy to capture common behaviors among the label-wise graph walk agents. The centralized policy serves as a global regularization to organize collaborative policy update of each agent. Thanks to the algorithmic design, we incorporate   i) the predicative relation between the structural attributes of the graph and labels and ii) the correlation between label-wise classification tasks in the proposed MLGW model. They are the two main pillars of a successful multi-label classification algorithm. Therefore, in contrast with other state-of-the-art approaches, we witness significantly higher classification accuracy and better graph exploration produced by our method in the comprehensive comparative study. 
	
	From the encouraging results of the MLGW model, there are several new directions for future work. One interesting direction would be to let the model decide when to terminate the walk, thereby eliminating the need for a fixed walk length $T$. Finally, we would want to make the model more efficient by reducing the number of computations (i.e., complexity).
	
	\section{Acknowledgment}
	This work is supported by the King Abdullah University of Science and Technology (KAUST), Saudi Arabia
	
	\bibliographystyle{IEEEtran}
	\bibliography{ijcai19}

\begin{thebibliography}{10}
\providecommand{\url}[1]{#1}
\csname url@samestyle\endcsname
\providecommand{\newblock}{\relax}
\providecommand{\bibinfo}[2]{#2}
\providecommand{\BIBentrySTDinterwordspacing}{\spaceskip=0pt\relax}
\providecommand{\BIBentryALTinterwordstretchfactor}{4}
\providecommand{\BIBentryALTinterwordspacing}{\spaceskip=\fontdimen2\font plus
\BIBentryALTinterwordstretchfactor\fontdimen3\font minus
  \fontdimen4\font\relax}
\providecommand{\BIBforeignlanguage}[2]{{%
\expandafter\ifx\csname l@#1\endcsname\relax
\typeout{** WARNING: IEEEtran.bst: No hyphenation pattern has been}%
\typeout{** loaded for the language `#1'. Using the pattern for}%
\typeout{** the default language instead.}%
\else
\language=\csname l@#1\endcsname
\fi
#2}}
\providecommand{\BIBdecl}{\relax}
\BIBdecl

\bibitem{hamilton2017inductive}
W.~Hamilton, Z.~Ying, and J.~Leskovec, ``Inductive representation learning on
  large graphs,'' in \emph{NIPS}, 2017, pp. 1024--1034.

\bibitem{gcn}
T.~N. Kipf and M.~Welling, ``Semi-supervised classification with graph
  convolutional networks,'' \emph{arXiv preprint arXiv:1609.02907}, 2016.

\bibitem{yang2016revisiting}
Z.~Yang, W.~W. Cohen, and R.~Salakhutdinov, ``Revisiting semi-supervised
  learning with graph embeddings,'' \emph{arXiv preprint arXiv:1603.08861},
  2016.

\bibitem{lee2018attention}
J.~B. Lee, R.~A. Rossi, S.~Kim, N.~K. Ahmed, and E.~Koh, ``Attention models in
  graphs: A survey,'' \emph{arXiv preprint arXiv:1807.07984}, 2018.

\bibitem{wu2014semi}
Q.~Wu, Y.~Ye, S.-S. Ho, and S.~Zhou, ``Semi-supervised multi-label collective
  classification ensemble for functional genomics,'' \emph{BMC genomics},
  vol.~15, no.~9, p. S17, 2014.

\bibitem{zha2009graph}
Z.-J. Zha, T.~Mei, J.~Wang, Z.~Wang, and X.-S. Hua, ``Graph-based
  semi-supervised learning with multiple labels,'' \emph{Journal of Visual
  Communication and Image Representation}, vol.~20, no.~2, pp. 97--103, 2009.

\bibitem{grover2016node2vec}
A.~Grover and J.~Leskovec, ``node2vec: Scalable feature learning for
  networks,'' in \emph{SIGKDD}, 2016, pp. 855--864.

\bibitem{perozzi2014deepwalk}
B.~Perozzi, R.~Al-Rfou, and S.~Skiena, ``Deepwalk: Online learning of social
  representations,'' in \emph{SIGKDD}, 2014, pp. 701--710.

\bibitem{tang2015line}
J.~Tang, M.~Qu, M.~Wang, M.~Zhang, J.~Yan, and Q.~Mei, ``Line: Large-scale
  information network embedding,'' in \emph{World Wide Web}, 2015, pp.
  1067--1077.

\bibitem{nandanwar2016structural}
S.~Nandanwar and M.~N. Murty, ``Structural neighborhood based classification of
  nodes in a network,'' in \emph{SIGKDD}.\hskip 1em plus 0.5em minus
  0.4em\relax ACM, 2016, pp. 1085--1094.

\bibitem{gao2018deep}
H.~Gao and H.~Huang, ``Deep attributed network embedding.'' in \emph{IJCAI},
  2018, pp. 3364--3370.

\bibitem{yang2015network}
C.~Yang, Z.~Liu, D.~Zhao, M.~Sun, and E.~Y. Chang, ``Network representation
  learning with rich text information.'' in \emph{IJCAI}, 2015, pp. 2111--2117.

\bibitem{chen2017hierarchical}
W.~Chen, J.~Wang, Z.~Jiang, Y.~Zhang, and X.~Li, ``Hierarchical mixed neural
  network for joint representation learning of social-attribute network,'' in
  \emph{PA-KDD}.\hskip 1em plus 0.5em minus 0.4em\relax Springer, 2017, pp.
  238--250.

\bibitem{akujuobi2018mining}
U.~Akujuobi, K.~Sun, and X.~Zhang, ``Mining top-k popular datasets via a deep
  generative model,'' in \emph{2018 IEEE Big Data}.\hskip 1em plus 0.5em minus
  0.4em\relax IEEE, 2018, pp. 584--593.

\bibitem{zhuang2018dual}
C.~Zhuang and Q.~Ma, ``Dual graph convolutional networks for graph-based
  semi-supervised classification,'' in \emph{World Wide Web}, 2018, pp.
  499--508.

\bibitem{velickovic2017graph}
P.~Velickovic, G.~Cucurull, A.~Casanova, A.~Romero, P.~Lio, and Y.~Bengio,
  ``Graph attention networks,'' \emph{arXiv preprint arXiv:1710.10903}, vol.~1,
  no.~2, 2017.

\bibitem{LANE-WSDM17}
X.~Huang, J.~Li, and X.~Hu, ``Label informed attributed network embedding,'' in
  \emph{WSDM}, 2017, pp. 731--739.

\bibitem{xiong2017deeppath}
W.~Xiong, T.~Hoang, and W.~Y. Wang, ``Deeppath: A reinforcement learning method
  for knowledge graph reasoning,'' \emph{arXiv preprint arXiv:1707.06690},
  2017.

\bibitem{hoshen2017vain}
Y.~Hoshen, ``Vain: Attentional multi-agent predictive modeling,'' in
  \emph{NIPS}, 2017, pp. 2701--2711.

\bibitem{jiang2018graph}
J.~Jiang, C.~Dun, and Z.~Lu, ``Graph convolutional reinforcement learning for
  multi-agent cooperation,'' \emph{arXiv preprint arXiv:1810.09202}, 2018.

\bibitem{lee2017deep}
J.~B. {Lee}, R.~A. {Rossi}, and X.~{Kong}, ``Deep graph attention model.''
  \emph{arXiv preprint arXiv:1709.06075}, 2017.

\bibitem{METaylor2017AAAI}
S.E.Bsat, H.Bou-Ammar, and M.E.Taylor, ``Scalable multitask policy gradient
  reinforcement learning,'' in \emph{AAAI}, ser. AAAI'17, 2017, pp. 1847--1853.

\bibitem{Tutunov2018NIPS}
R.Tutunov, D.Kim, and H.Bou-Ammar, ``Distributed multitask reinforcement
  learning with quadratic convergence,'' in \emph{Neural Information Processing
  Systems}, ser. NIPS'18, 2018, pp. 8921--8930.

\bibitem{Calandriello2014NIPS}
D.Calandriello, A.Lazaric, and M.Restelli, ``Sparse multi-task reinforcement
  learning,'' in \emph{Neural Information Processing Systems}, 2014.

\bibitem{Mnih2016AsynchronousMF}
V.~Mnih, A.~P. Badia, M.~Mirza, A.~Graves, T.~P. Lillicrap, T.~Harley,
  D.~Silver, and K.~Kavukcuoglu, ``Asynchronous methods for deep reinforcement
  learning,'' in \emph{International Conference on Machine Learning}, 2016.

\bibitem{Espeholt2018}
L.Espeholt, H.Soyer, R.Munos, K.Simonyan, V.Mnih, T.Ward, Y.Doron, V.Firoiu,
  T.Harley, I.Dunning, S.Legg, and K.Kavukcuoglu, ``Impala: Scalable
  distributed deep-rl with importance weighted actor-learner achitectures,'' in
  \emph{International Conference on Machine Learning}, 2018, pp. 570--586.

\bibitem{Hessel2019MultitaskDR}
M.~Hessel, H.~Soyer, L.~Espeholt, W.~Czarnecki, S.~Schmitt, and H.~P. van
  Hasselt, ``Multi-task deep reinforcement learning with popart,'' in
  \emph{AAAI}, 2019.

\bibitem{Teh2017DistralRM}
Y.~W. Teh, V.~Bapst, W.~Czarnecki, J.~Quan, J.~Kirkpatrick, R.~Hadsell,
  N.~Heess, and R.~Pascanu, ``Distral: Robust multitask reinforcement
  learning,'' in \emph{Neural Information Processing Systems}, 2017.

\bibitem{cho2014learning}
K.~Cho, B.~van Merrienboer, C.~Gulcehre, D.~Bahdanau, F.~Bougares, H.~Schwenk,
  and Y.~Bengio, ``Learning phrase representations using rnn encoder--decoder
  for statistical machine translation,'' in \emph{EMNLP}, 2014, pp. 1724--1734.

\bibitem{Shen2018MWalkLT}
Y.~Shen, J.~Chen, P.-S. Huang, Y.~Guo, and J.~Gao, ``M-walk: Learning to walk
  over graphs using monte carlo tree search,'' in \emph{NeurIPS}, 2018.

\bibitem{Yu:2014}
H.-F. Yu, P.~Jain, P.~Kar, and I.~S.Dhillon, ``Large-scale multi-label learning
  with missing labels,'' in \emph{Proceedings of the 31st International
  Conference on International Conference on Machine Learning - Volume 32}, ser.
  ICML'14, 2014, pp. I--593--I--601.

\bibitem{williams1992simple}
R.~J. Williams, ``Simple statistical gradient-following algorithms for
  connectionist reinforcement learning,'' \emph{Machine learning}, vol.~8, no.
  3-4, pp. 229--256, 1992.

\bibitem{mnih2014recurrent}
V.~Mnih, N.~Heess, A.~Graves \emph{et~al.}, ``Recurrent models of visual
  attention,'' in \emph{NIPS}, 2014, pp. 2204--2212.

\bibitem{ji2010graph}
M.~Ji, Y.~Sun, M.~Danilevsky, J.~Han, and J.~Gao, ``Graph regularized
  transductive classification on heterogeneous information networks,'' in
  \emph{Joint European Conference on Machine Learning and Knowledge Discovery
  in Databases}.\hskip 1em plus 0.5em minus 0.4em\relax Springer, 2010, pp.
  570--586.

\bibitem{zhang2014review}
M.-L. Zhang and Z.-H. Zhou, ``A review on multi-label learning algorithms,''
  \emph{TKDE}, vol.~26, no.~8, pp. 1819--1837, 2014.

\bibitem{5567103}
G.~Tsoumakas, I.~Katakis, and I.~Vlahavas, ``Random k-labelsets for multilabel
  classification,'' \emph{IEEE Transactions on Knowledge and Data Engineering},
  vol.~23, no.~7, pp. 1079--1089, July 2011.

\bibitem{zhang2007ml}
M.-L. Zhang and Z.-H. Zhou, ``Ml-knn: A lazy learning approach to multi-label
  learning,'' \emph{Pattern recognition}, vol.~40, no.~7, pp. 2038--2048, 2007.

\bibitem{szymanski2018lnemlc}
P.~Szyma{\'n}ski, T.~Kajdanowicz, and N.~Chawla, ``Lnemlc: Label network
  embeddings for multi-label classifiation,'' \emph{arXiv preprint
  arXiv:1812.02956}, 2018.

\bibitem{7395756}
F.~Benites and E.~Sapozhnikova, ``Haram: A hierarchical aram neural network for
  large-scale text classification,'' in \emph{ICDMW}, 2015, pp. 847--854.

\bibitem{sechidis2011stratification}
K.~Sechidis, G.~Tsoumakas, and I.~Vlahavas, ``On the stratification of
  multi-label data,'' in \emph{Joint European Conference on Machine Learning
  and Knowledge Discovery in Databases}.\hskip 1em plus 0.5em minus 0.4em\relax
  Springer, 2011, pp. 145--158.

\end{thebibliography}
	
\end{document}